\documentclass{svproc}
\usepackage{url}

\usepackage{amsmath} % assumes amsmath package installed
\usepackage{amssymb}  % assumes amsmath package installed
\usepackage{soul, color}
\usepackage{amsfonts}
\usepackage{graphicx}
\usepackage{breakcites}

\begin{document}
\mainmatter              % start of a contribution
\title{Model Based Control of Commercial-Off-The Shelf (COTS) Unmanned Rotorcraft for Brick Wall Construction}
\titlerunning{Model Based Control of COTS Rotorcraft}  % abbreviated title (for running head)
\author{Nithya Sridhar\inst{1} \and N. Sai Abhinay\inst{1},
B. Chaithanya Krishna\inst{1}\and Shubhankar Shobhit\inst{1} \and Kaushik Das\inst{1} \and Debasish Ghose\inst{2}}
%\inst{1}
\authorrunning{Nithya Sridhar et al.} % abbreviated author list (for running head)
%Craig Chambers, Kim B. Bruce, and Elisa Bertino}
%
\institute{ ${}^1$ Tata Consultancy Services Innovation Labs, Bangalore -560066, India,\\
\email{\{sridhar.nithya, ns.abhinay, c.bodduluri, s.shobhit, kaushik.da\}@tcs.com},\\ %WWW home page:
\texttt{}
${}^2$Indian Institute of Science, Bangalore -560012, India\\
\email{dghose@iisc.ac.in}
}
%\and
%Universit\'{e} de Paris-Sud, 

%Laboratoire d'Analyse Num\'{e}rique, B\^{a}timent 425,\\
%F-91405 Orsay Cedex, France

\maketitle              % typeset the title of the contribution

\begin{abstract}
This work proposes a systematic framework for modelling and controller design of a  Commercial-Off-The Shelf (COTS) unmanned rotorcraft using control theory and principles, for brick wall construction. With point to point navigation as the primary application, command velocities in the three axes of the Unmanned Aerial Vehicle (UAV) are considered as inputs of the system while its actual velocities are system outputs. Using the sine and step {response} data acquired from a Hardware-in-Loop (HiL) test simulator, the considered system was modelled in individual axes with the help of the {proposed framework}. {This model was employed for controller design where a sliding mode controller was chosen to satisfy certain requirements of the application like robustness, flexibility and accuracy.} The model was validated using step response data and produced a deviation of only 9\%. Finally, the controller results from field test showed fine control up to $\pm8$ cms accuracy. Sliding Mode Control {(SMC) was also compared with a linear controller derived from iterative experimentations and seen to perform better than the latter in terms of accuracy, and robustness to parametric variations and wind disturbances.}
% We would like to encourage you to list your keywords within
% the abstract section using the \keywords{...} command.
\end{abstract}

\section{Introduction}

The new age of unmanned systems has been witnessing rapid expansion in the Unmanned Aerial Vehicles' (UAVs) application horizon besides its current use in surveillance, agriculture, defence, material delivery, etc. With a further futuristic vision, Mohamed Bin Zayed International Robotics Challenge (MBZIRC 2020) had framed three challenges for the participating teams. One of them consisted of a task where the UAVs are to be employed to build brick walls using special bricks specifically made for the competition. The sequence of tasks involved were multi-fold 1) Detection of brick piles and construction zone 2) Picking of different coloured bricks of different lengths and weight, with magnetic aids 3) Placing of different coloured bricks in the construction zone as per the given pattern.

%\begin{enumerate}
%\item Detection of brick piles and construction zone
%\item Picking of different coloured bricks of different lengths and weight, with magnetic aids
%\item Placing of different coloured bricks in the construction zone as per the given pattern.
%\end{enumerate}

This work has made use of unmanned rotorcraft as the UAV for the given competetion. It is important to note that UAV navigation was an integral part in all the mentioned tasks. This paper addresses the problem of point to point navigation for brick wall construction and attempts to fuse the benefits of two contrasting approaches. One of which involved a heuristic experiment based methodology and the other was analytical. This work considers the practical constraints and challenges involved in the task, and also uses control theory extensively in the framework. The results of the proposed framework are compared against that of its heuristic counterpart with the help of experimental results.

The paper begins with explaining the tasks in the competition and the challenges involved in it. Further, section~\ref{sec2} discusses the importance of UAV modelling in controller design, along with some challenges in COTS rotorcraft modelling. Section~\ref{sec3} discusses two contrasting control approaches with its features. Further, section~\ref{sec4} and~\ref{sec5} present the modelling framework and controller design process in detail followed by field results in section~\ref{sec6}.
\subsection{Challenges in Brick Wall Construction}
\begin{figure}[h!]
\center
{{\includegraphics[trim=0cm 0cm 0cm 0cm, clip=true, totalheight=0.35\textheight,]{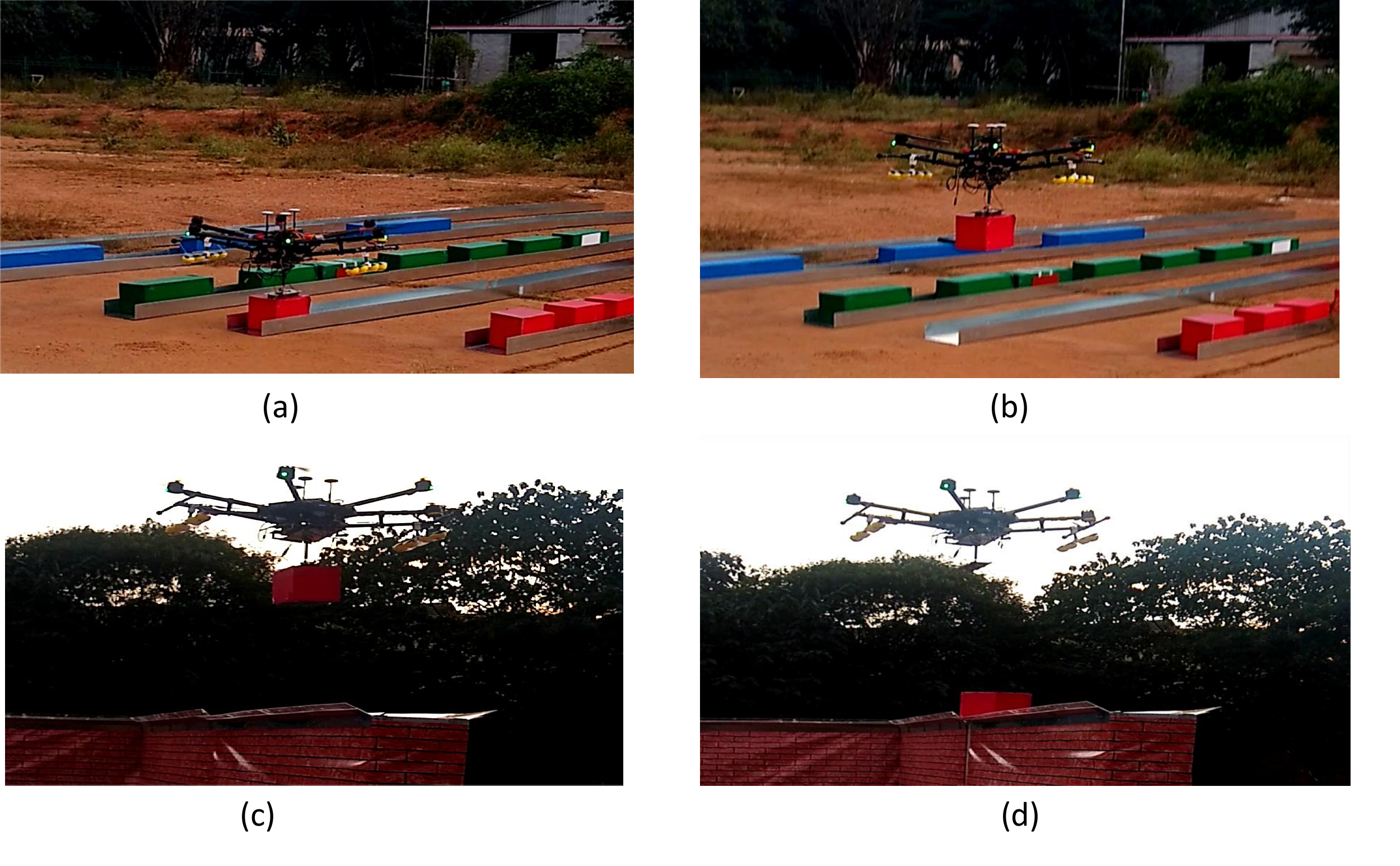}}}
\caption{{Field test with SMC (a) Brick picking attempted (b) Brick is lifted in air (c) Brick placing attempted (d) Brick placed in the channel}} \label{fig:pick}
\end{figure}

{Brick wall construction involved picking of bricks from a brick pile zone where different coloured bricks were available in a particular order similar to that shown in} Figures~\ref{fig:pick} (a) \& (b). {The picking action involved brick detection using vision feedback, accurate position control and gripper actuation for attachment. After lifting the brick in air, the UAV would navigate to the given construction zone similar to that shown in} Figures~\ref{fig:pick} (c) \& (d). {This involved fast but reliable navigation of the UAV without detaching the brick on the way. Further, the placing action involved placing area detection in the construction zone using vision feedback, fine position control and gripper detachment action. Also, the width of the bricks was 20 cms making finer UAV control a necessity. Hence, specific challenges involved with this application were:}
\begin{enumerate}
    \item {During picking and placing actions, accurate and finer control was a challenge given the small brick width of 20 cms. With this in mind and wide gripper design, the controller variation was expected to be not more than $\pm8$ cms.}
    \item {There was a maximum mass variation of 2 kg for the UAV while completing one set of brick picking and placing operation. This would lead to system parametric variations from control theory's viewpoint and had to be handled.}
    \item {Since the tasks were carried out in an open arena, wind disturbance was a huge concern.}
    \item {During the navigation from brick pile zone to construction site, the UAV controller should be reliable enough not to drop the brick on the way. However, in the case where it goes to pick the brick from the construction site, it has to be faster and aggressive in order to save time.}
\end{enumerate}
{Hence, for this particular application of point to point navigation, a high fidelity robust controller was required which could also provide flexibility to adapt to different task requirements and finer control to achieve the required accuracy.}
\section{Commercial UAV Modelling}\label{sec2}
Modelling of UAVs with multirotor configuration has been addressed in literature using physics-based models, empirical models, data-driven models, etc. For the physics-based models, identification of system parameters may require certain amount of testing and test facilities, which might make the identification process tedious~\cite{hoffer2014survey}. Other estimation techniques like least square estimation~\cite{gremillion2010system}, Kalman filtering techniques~\cite{ekf}~\cite{ukf} require prior information of the modelling structure.
{Moreover, these modelling frameworks do not consists of an unknown in-built stabilizer that is present in commercial UAVs.}

\begin{figure}[h]
\center
{{\includegraphics[trim=1cm 5cm 1cm 2cm, clip=true, totalheight=0.35\textheight,]{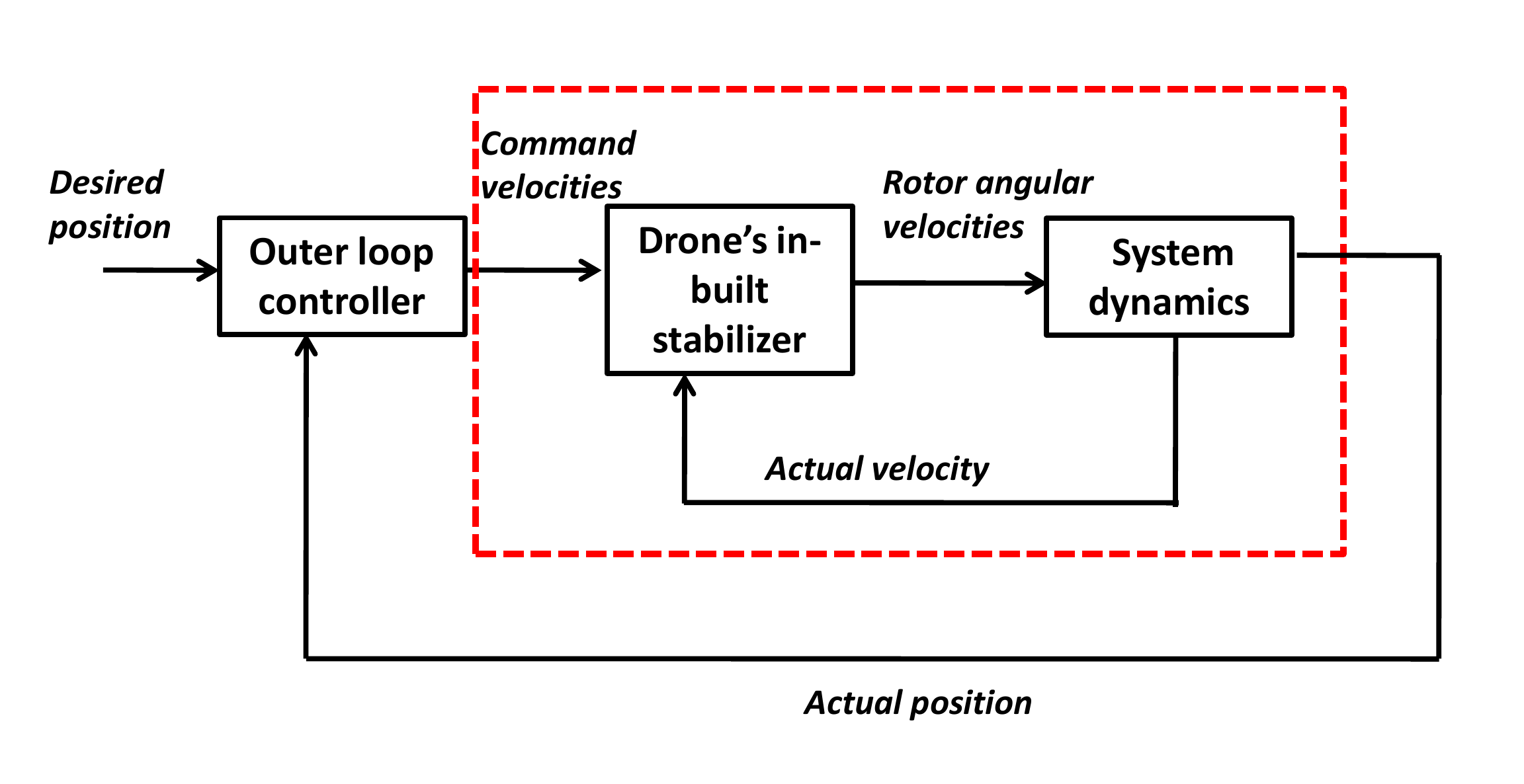}}}
\caption{{System block diagram}} \label{fig:blockdiag}
\end{figure}

Figure~\ref{fig:blockdiag} shows the block diagram of the considered UAV system with the in-built stabilizer. This stabilizer acts as an inner loop controller for the under-actuated UAV system to maintain system stability. In commercial UAVs, information about the stabilizer architecture is not available. Hence, it is imperative to adopt a more pragmatic approach while attempting to model commercial UAVs with unknown stabilizer details. This can be carried out using experimental data and empirical methods. It is also because of this lack of information that UAV application developers generally choose to go for heuristic approach for controller design. This may consist of a linear controller with one or more of Proportional, Derivative and/or Integral actions. This approach may not yield an optimal performance.

However, if the system dynamics are captured, controller design based on this would facilitate finer and accurate control within the drone's actuation limits. The proposed framework consists of this analytical-based approach considering practical constraints, which is believed to provide efficient control.

\section{Position Control Design: Heuristic Approach vs Analytical Approach}\label{sec3}
As mentioned in the previous section, commercially available UAVs consists of an in-built stabilizer, whose model is unknown to the user. It is also well known that model-based controller have good accuracy and facilitate superior control. However, with the missing piece of information, designing a model-based controller becomes a challenge. Hence, heuristic approaches are largely prevalent among the UAV community, which involves controller design from experiments. By and large, a linear controller is preferred with few or all of the following control actions: Proportional, Integral and Derivative. The choice of control actions and controller parameter values are obtained from iterative experimentations. This approach may address the control problem partially and may not satisfy performance criteria all the time. With its conservative approach, the controller performance may not be improved beyond a certain level. Some of the features of this approach are:
\begin{enumerate}
\item Time required for controller implementation is minimal
\item Low complexity in controller design
\item Does not require the model of in-built stabilizer
\item Produces sub-optimal performance.
\item Only target and actual values of the control variable are required for implementation
\end{enumerate}

On the other hand, analytical approaches~\cite{najm2019nonlinear}~\cite{kamel2017dynamic} require the system model and its parameters for controller design. Consideration of system dynamics in the control design process improves the efficacy of the controller, provides flexibility to satisfy the desired performance criteria. Control theory is genarally used in the design  and considerably less experimentation is required for design. With good model accuracy, close correlation between theoretical and experimental responses can be obtained. Some features of this approach are:
\begin{enumerate}
\item Time required for controller implementation depends on the type of controller
\item Simplicity of controller design depends on the type of controller
\item Requires the model of in-built stabilizer
\item Capable of producing superior performance
\item More signal variables may be required other than target and actual control variables
\end{enumerate}

A judicious mix of both these approaches would aid the controller development stage and is believed to provide acceptable performance.

\section{COTS Rotorcraft Modelling Framework}\label{sec4}
With the well known advantages of model-based controller, it is imperative to obtain a mathematical model of the commercial rotorcraft. Literature has discussed numerous techniques to model UAV, however the presence of an in-built stabilizer and the absence of its model information is a huge challenge. This was addressed by adopting certain features of a framework proposed for vehicle brake system in literature~\cite{sridhar2017model}. The stabilizer and the UAV  were considered as one single unit for modelling. Since the application was point to point navigation that involved position control, velocity in each axis was taken as the control command variable. Also with the Inertial Measurement Unit (IMU), the actual velocity signals of the drone in all the three axes of translation was available for external use. Hence, the system for modelling  considered command velocities as input and actual velocities as output for individual axes. 

The Hardware-in-Loop simulator platform, provided by the manufacturer was used to acquire data for modelling. Firstly, sine responses were obtained till 15 rad/s, which was used to check for linearity and time in-variance properties (LTI). Later, the magnitude plots in the frequency domain were used to obtain the model structure and its parameters. Further, experimental step response data was compared with the simulated response of the fitted model to check its validity.   
\subsection{Frequency Domain Analysis}
\begin{figure}[htb!]
\center
{{\includegraphics[trim=2cm 0cm 2cm 0cm, clip=true, totalheight=0.3\textheight,]{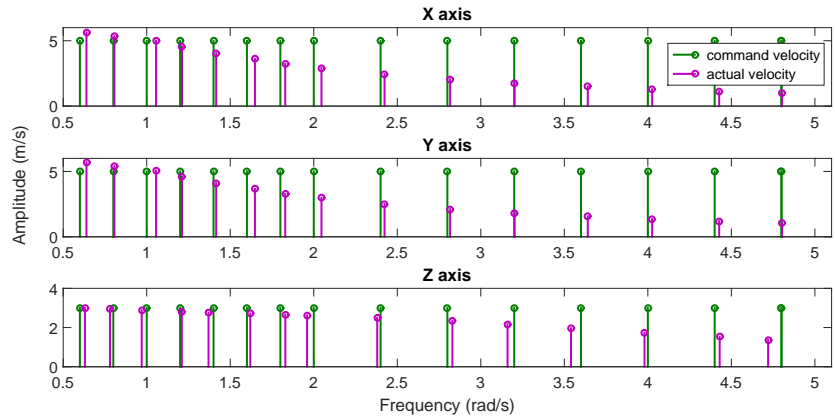}}}
\caption{{Spectral analysis}} \label{fig:spectrum}
\end{figure}

\begin{figure}[bth!]
\center
{{\includegraphics[trim=2cm 0cm 10cm 0cm, clip=true, totalheight=0.35\textheight,]{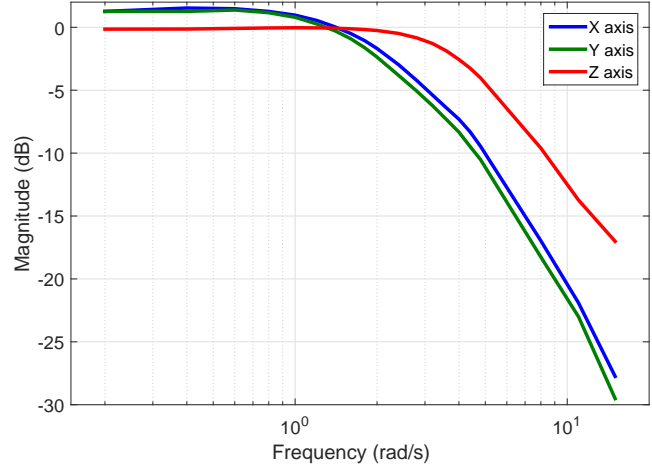}}}
\caption{{Magnitude plot}} \label{fig:bode}
\end{figure}

The sine response data obtained using HiL simulator was analysed to check if the system could be approximated as an LTI system. The peak frequency components of the actual velocities were plotted against that of the input waveform as shown in Fig.~\ref{fig:spectrum}. It was observed that the output frequency of the peak component was very close to input frequency value with a slight average deviation in the order of $10^{-2}$. This ascertained that the system can be approximated as LTI. It was also observed that the responses of X and Y axes were very close. Hence, it was decided to obtain a single model for both X and Y axes.
Frequency domain analysis was carried out using sine sweep data obtained for the frequency range of [0.4, 15] rad/s for different amplitude values ranging from [0.5, 5] m/s for X and Y axes, and that of [0,3] m/s for Z axis. Magnitude plots were obtained using the acquired data, of which one sample for each axis is shown in Fig.~\ref{fig:bode}. It was seen that X and Y axes characteristics were in close agreement supporting the previous inference from spectral analysis. 
With nearly zero slope in the lower asymptote region, it was clear that the system did not have any zeros in the transfer function. In the high frequency asymptote region, the slopes of X, Y and Z axes were approximately -20 dB/decade, indicative of system's relative order to be one. Since, the system has no zeros, with relative order one, it has a pole and takes the form of
\begin{equation}
G(s)=\frac{K}{1+\tau s}, \label{eq:sys}
\end{equation}
in the Laplace domain, where $K$ is the steady state gain and $\tau$ is the system time constant. Steady state gain $K$ was obtained from the low frequency asymptote region using,
\begin{equation}
K=10^{\frac{M(\omega_l)}{20}},
\end{equation}
where $M(\omega_l)$ is the magnitude of the system at the lowest frequency considered ($\omega_l=0.4$ rad/s).
With the family of curves obtained for the three axes systems, the range of time constant was obtained using,
\begin{equation}
\tau=\frac{1}{2\pi f_c},
\end{equation}
where $f_c$ is the experimental cut-off frequency.
The ranges of time constant for X and Z axes were [0.5, 0.9] s and [0.2 0.4] s respectively.
Using the time constant ranges, model structure and sine response data, the time constant $\tau$ was estimated using simple least square estimation technique~\cite{gremillion2010system}.

The transfer functions for X, Y, and Z axes are,
\begin{equation}
G_x(s)=G_y(s)=\frac{1.16}{1+0.75s}, \label{eq:sys_x}
\end{equation}
and,
\begin{equation}
G_z(s)=\frac{0.98}{1+0.30s}, \label{eq:sys_z}
\end{equation}
respectively.

\subsection{Time Domain Analysis}
Step response data was acquired for amplitudes ranging from 1 m/s to 5 m/s in steps of 1 m/s. The responses were captured for each axis individually and was compared against the simulated step response of the models given in Equations~\ref{eq:sys_x},~\ref{eq:sys_y} and ~\ref{eq:sys_z}. The deviation between these responses were quatified using Mean Absolute Percentage Deviation (MAPD) defined as,
\begin{equation}
MAPD=\left[\frac{1}{N}{\sum\limits_{k=0}^N\Bigg|{\frac{g_{exp}(k)-g_{sim}(k)}{g_{exp}(k)}}\Bigg|}\right]100,\label{eq:MAPD}
\end{equation}
where $g_{exp}$ is the experimental data point, $g_{sim}$ is the simulated data point at time interval $k$ with $N$ number of data points.
\begin{figure}[h!]
\center
{{\includegraphics[trim=2cm 2cm 12cm 0cm, clip=true, totalheight=0.35\textheight,]{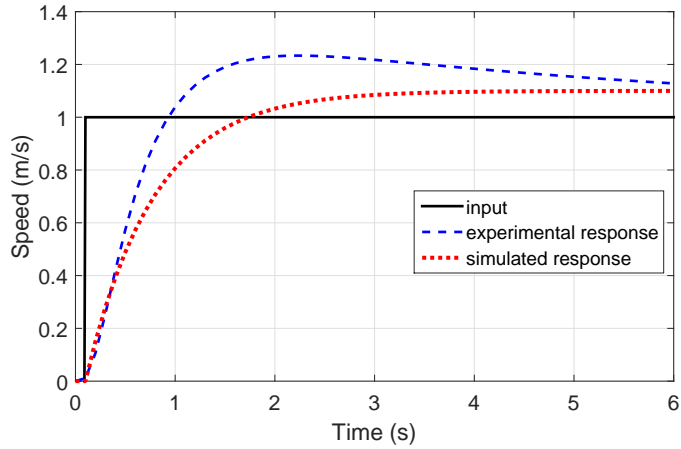}}}
\caption{{Step response- X axis}} \label{fig:time_x}
\end{figure}

\begin{figure}[h!]
\center
{{\includegraphics[trim=2cm 2cm 12cm 0cm, clip=true, totalheight=0.35\textheight,]{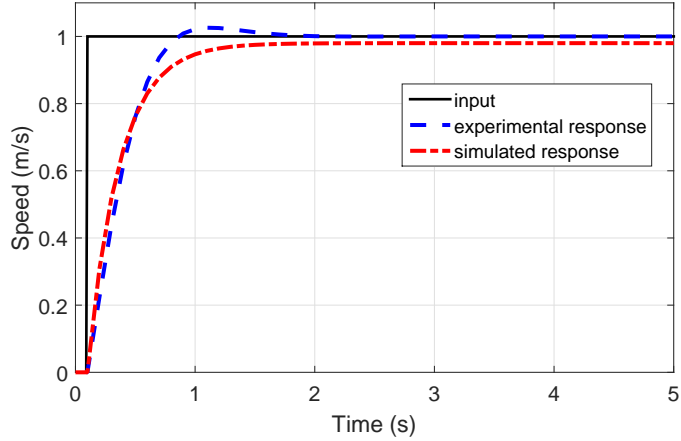}}}
\caption{{Step response- Z axis}} \label{fig:time_z}
\end{figure}
Figures~\ref{fig:time_x} and~\ref{fig:time_z} show the comparison of simulated and experimental step response. The experimental response shows a slight overshoot which is not captured in the model, thus producing a maximum MAPD of 16\% for X axis and 12\%  for Z axis. However, the average MAPD was approximately only 9\% and the estimated model was used for model-based controller design with this small deviation.

\section{Sliding Mode Controller for Point to Point Navigation}\label{sec5}
\begin{figure}[h]
\center
{{\includegraphics[trim=2cm 5cm 1cm 2cm, clip=true, totalheight=0.3\textheight,]{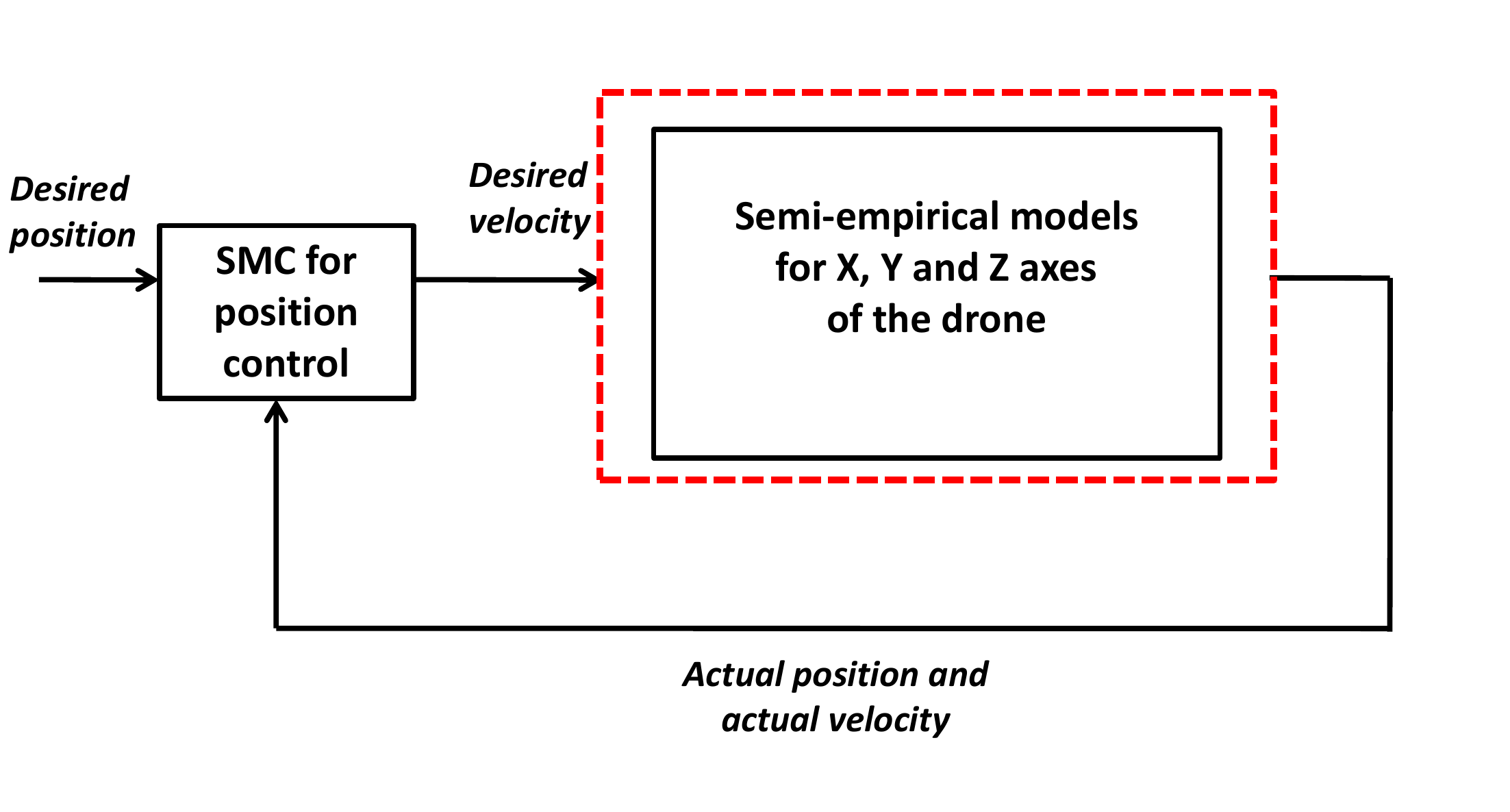}}}
\caption{{System block diagram}} \label{fig:smc_control}
\end{figure}

Model-based controller development was attempted with the use of the estimated models. Many control techniques such as sliding mode control, model predictive control~\cite{kamel2017model}, fuzzy logic control~\cite{lv2018fuzzy}, etc. are discussed in literature. However, a recent study~\cite{nasr2018comparitive} showed that sliding mode control technique was the best when compared with other popular techniques for trajectory tracking. Also, from competition viewpoint, the following factors influenced  controller's choice:
\begin{enumerate}
\item With a maximum MAPD of 16\% in the modelling process, it was imperative to choose an inherently robust controller that can handle unmodelled dynamics of the system.
\item Since the mass of the drone was expected to vary during the picking and placing tasks of brick, robustness was required in terms of system parametric variations.
\item Given the variable nature of wind, robustness was a much required property.
\end{enumerate}
Considering all these factors, it was decided to implement sliding mode control, which was known to be robust against parametric variations and external disturbances. Figure~\ref{fig:smc_control} shows the control system design with sliding mode controller for position control.
Equation~\ref{eq:sys} can be written in state space form as,
\begin{equation}
\dot{y(t)}=Ay(t)+Bu(t)\label{eq:statespace}
\end{equation}
where $y(t)$ is output velocity, $u(t)$ is the command velocity and, $A=\frac{-1}{\tau}$, $B=\frac{K}{\tau}$ are system parameters for a single axis system. Position of the drone is given by,
\begin{equation}
p(t)=\int y(t)dt \label{eq:pos}
\end{equation}
assuming the drone is starting at zero position. Since position of the drone was to be controlled by the outer loop controller, the aim was to reduce the position error, given by,
\begin{equation}
e(t)=p(t)-p_d(t)\label{eq:error}
\end{equation}
where $p(t)$ is the actual position and $p_d(t)$ is the desired position.
The sliding surface was defined as,
\begin{equation}
s(t)=\lambda e(t)+\dot{e}(t)\label{eq:ss}
\end{equation}
Differentiating Equation~\ref{eq:ss} and substituting Equations~\ref{eq:error},~\ref{eq:pos} and~\ref{eq:statespace},
\begin{equation}
\dot{s}(t)=\lambda y(t) + A y(t) + B u(t),\label{eq:s_dot}
\end{equation}
where the rate of change of desired position was assumed to be zero.
The reaching law for SMC was considered as,
\begin{equation}
\dot{s}(t)=-Ksign(s)-Qs\label{eq:CPRRL}
\end{equation}
where $K$ is the gain of constant rate reaching law and $Q$ is the proportional gain of the second term considered in the law. {This reaching law was particularly chosen with proportional term so that the aggressiveness of the UAV can be varied for different task requirements.}
From Equations~\ref{eq:s_dot} and~\ref{eq:CPRRL}, the control input $u(t)$ is derived as,
\begin{equation}
u(t)=\frac{-Ksign(s)-Qs-Ay(t)-\lambda y(t)}{B}\label{eq:control}
\end{equation}
The derived control input was given as command velocity for position control for navigation.
\section{Field Results}\label{sec6}
The drone was tested in the field with the model-based SMC controller as well as the linear controller obtained from heuristic approach. The linear controller consisted of proportional and derivative actions (PD). The drone was given a target point in the three dimensional space starting at the origin. It was programmed to land when a desired accuracy of $\pm8$ cm was achieved and maintained in each axis for a time period of 5 seconds. It was also ensured that safe saturation limits were implemented during drone actuation in trials.
\begin{figure}[h!]
\center
{{\includegraphics[trim=2cm 0cm 2cm 0cm, clip=true, totalheight=0.32\textheight,]{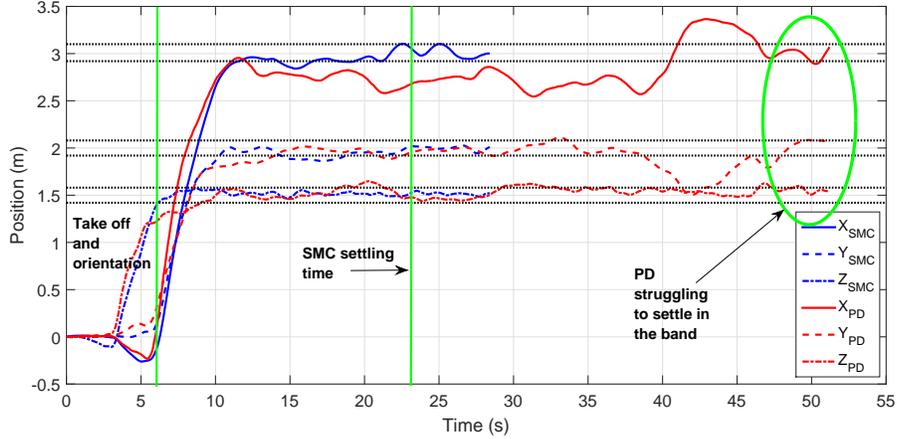}}}
\caption{{Comparison of PD and SMC performance}} \label{fig:pd_smc}
\end{figure}

Figure~\ref{fig:pd_smc} shows the comparison of SMC and PD performance. Time period for the drone's take off action and orientation to East direction, is marked in the figure. Once the drone reached its minimum operational height, it started to move to the given target in the three dimensional space. This test was conducted in the presence of wind with an average wind speed of 5 km/h. In Fig.~\ref{fig:pd_smc}, SMC settled in the desired accuracy band at around 23 s, and landing occurred at around 28 s. The PD controller was not able to settle in the band of 8 cm due to wind disturbances even after 50 s. Another important observation was the large deviation in position for the PD controller in the presence of wind. 

For the competition, separate set of SMC's controller parameters were obtained for different bricks and the rise time was adjusted for laden and unladen conditions. SMC rise time was governed by the controller parameter, $Q$. This value was adjusted such that the drone did not behave aggressively while carrying the brick, owing to the possibility of dropping it. Similarly, in unladen condition, SMC was made aggressive by increasing $Q$, to save time.

From the results, it was inferred that SMC showed good robustness in windy conditions and was believed to have addressed the modelling inaccuracies and unmodelled dynamics. It also provided flexibility to adjust to different task requirements. 

\begin{figure}[h!]
\center
{{\includegraphics[trim=0cm 0cm 0cm 0cm, clip=true, totalheight=0.35\textheight,]{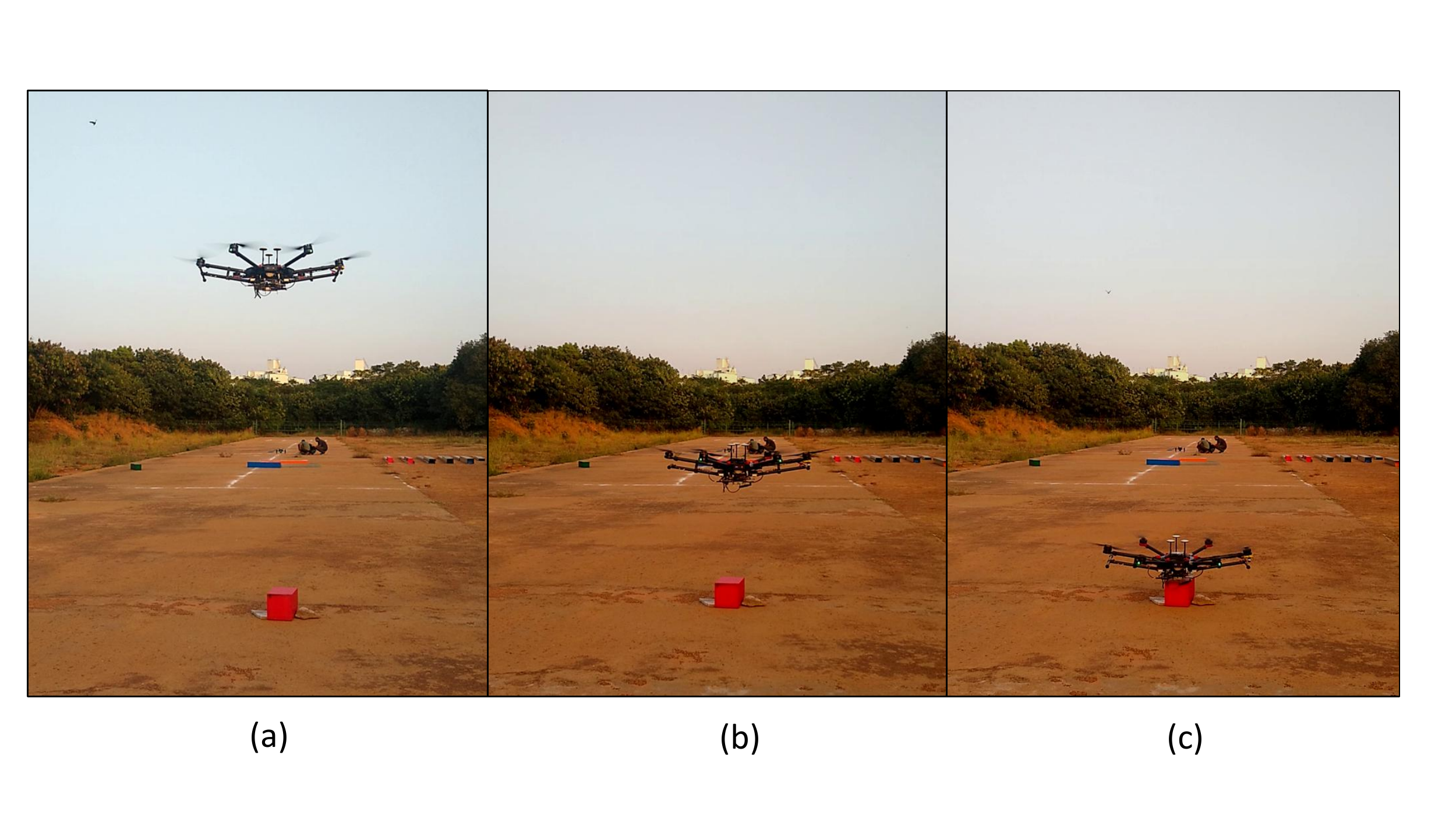}}}
\caption{{Field test with SMC (a) UAV searches for the brick (b) UAV detects the brick and approaches towards it (c) UAV attempts to pick the brick}} \label{fig:image}
\end{figure}

{Further, SMC controller was integrated with vision feedback algorithm and implemented for brick picking and placing.} Figure~\ref{fig:image} {shows an image of the drone attempting to pick up a brick using SMC, during field trials. SMC was able to carry out the required tasks with excellent accuracy, as seen from} Figure~\ref{fig:image}. {Details of integration with vision algorithm, controller parameter tuning for different scenarios and results will be included in the subsequent papers.}

\section{Conclusion}
A systematic framework was proposed for modelling of commercial UAV using the data acquired from its simulator. This method produced a first order model relating actual velocity of the drone to its command velocity for individual axis of translatory motion, with an average deviation of 9\% and a maximum deviation of 16\%. In order to account for the unmodelled dynamics, wind disturbances and parametric variations in the drone mass and inertia, a sliding mode controller was used with constant rate reaching law accompanied by a proportional term. This controller was compared with conventional PD controller obtained through a heuristic approach and was seen to perform better than the latter in terms of accuracy and robustness. For a fixed accuracy band of $\pm8$ cm, SMC was able to maintain the position within the band, however PD failed to do it. Thus, this framework provided a pragmatic methodology with analytical backing to design a model based controller for the intended application, considering task related constraints and environmental conditions during the controller development process.

\bibliographystyle{spbasic}
\bibliography{ref}

\end{document}